\documentclass{ecai} 

\usepackage{latexsym}
\usepackage{amssymb}
\usepackage{amsmath}
\usepackage{amsthm}
\usepackage{booktabs}
\usepackage{enumitem}
\usepackage{graphicx}
\usepackage{color}
\usepackage{float}
\usepackage{placeins}
\usepackage{caption}
\usepackage{hyperref}
\usepackage{url}

\newcommand{\BibTeX}{B\kern-.05em{\sc i\kern-.025em b}\kern-.08em\TeX}

\begin{document}


\begin{frontmatter}




\title{A separability-based approach to quantifying generalization: which layer is best?}


\author[A]{\fnms{Luciano}~\snm{Dyballa}\footnote{Equal contribution.\\\ Code available at \url{https://github.com/dyballa/generalization}}\thanks{Corresponding Author. Email: luciano.dyballa@yale.edu.}}
\author[A]{\fnms{Evan}~\snm{Gerritz}\footnotemark\thanks{Corresponding Author. Email: evan.gerritz@yale.edu.}}
\author[A]{\fnms{Steven W.}~\snm{Zucker}}

\address[A]{Department of Computer Science, Yale University, New Haven, CT, USA}


\begin{abstract}
Generalization to unseen data remains poorly understood for deep learning classification and foundation models, especially in the open set scenario.  How can one assess the ability of networks to adapt to new or extended versions of their input space in the spirit of few-shot learning, out-of-distribution generalization, domain adaptation, and category discovery? Which layers of a network are likely to generalize best? We provide a new method for evaluating the capacity of networks to represent a sampled domain, regardless of whether the network has been trained on all classes in that domain. Our approach is the following: after fine-tuning state-of-the-art pre-trained models for visual classification on a particular domain, we assess their performance on data from related but distinct variations in that domain. Generalization power is quantified as a function of the latent embeddings of unseen data from intermediate layers for both unsupervised and supervised settings. Working throughout all stages of the network, we find that (i) high classification accuracy does not imply high generalizability; and (ii) deeper layers in a model do not always generalize the best, which has implications for pruning. Since the trends observed across datasets are largely consistent, we conclude that our approach reveals (a function of) the intrinsic capacity of the different layers of a model to generalize.
\end{abstract}

\end{frontmatter}

\section{Introduction}

\begin{figure*}[ht!]
    \centering
    \includegraphics[width=1\textwidth]{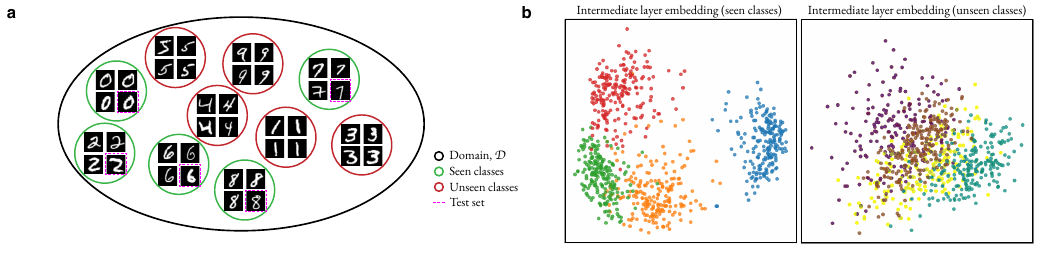}
    \caption{\textbf{Motivation for our approach.} (\textbf{a}) Example of a domain with an equivalence class structure. Some classes are used in training and model evaluation (\textit{seen}, in green) and the rest are not (\textit{unseen}, in red). (\textbf{b}) Typical example of the disparity between seen-class embeddings and unseen-class embeddings. Note the former are readily separable, but the latter are not, despite high test-set classification accuracy. This illustrates poor generalization. We formalize the representation's generalization quality by measures of separability for the unseen classes. Plots show embeddings of an intermediate layer output from VGG16 \cite{simonyan2014very}, visualized using PCA.}
    \label{fig:intro}
\end{figure*}

The extent to which a network represents a target domain is a key question for successful generalization. We work from the observation that an equivalence class structure underlies successful classification, and exploit this topology to develop a measure of generalizability based on separability (Figure \ref{fig:intro}). Our method examines the behavior of the intermediate layers on examples from classes missing in both the training and test sets, a problem confounding earlier attempts to quantify generalization to a different dataset with the same classes \citep{alain17understanding}. Importantly, our measure can be applied to any intermediate layer, allowing us to test the competing hypotheses that (i) early layers should capture basic, general features that are more easily translatable to other datasets, or that (ii) the deeper the representation is---and therefore closer to the final encoding/output layer---, the more ``useful'' it should be. Neither perspective, it turns out, is true.

\begin{figure}[ht]
    \centering
    \includegraphics[width=0.5\textwidth]{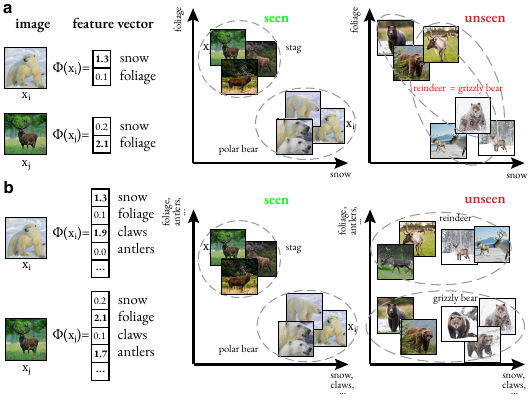}
    \caption{{\bf Certain rich feature spaces support clustering.}  (\textbf{a}) An impoverished model can classify stags vs. polar bears based on the background: foliage vs. snow, but fails on unseen examples. Note the lack of cluster separability (bears and reindeer are mixed due to their similar backgrounds). $\Phi(x)$ denotes the feature vector produced for the data point $x$. (\textbf{b}) A richer model also ``knows'' about antlers, claws, hooves, etc. and uses those to separate reindeer from grizzly bears, regardless of their background. Note the cluster separability.\\}
    \label{fig:cartoon}
\end{figure}

To empirically study a model's generalization capacity, we train it on a subset of the classes from a dataset (the \textit{seen} classes), and then investigate the model's behavior on the remaining classes (or \textit{unseen} classes).
The motivation for this approach is that the features learned for the seen classes should be used, only in different combinations, for representing the common features of the domain. Thus, unseen classes could be organized/separable within the same embedding space. To generalize well in this scenario, a network must have a sufficient number of neurons to represent a rich set of features that will also be found in the images from unseen classes; this idea is depicted in Figure \ref{fig:cartoon}. Hence, models that tend to learn more details (even those not necessarily useful for classification), in other words learning a richer representation of the features in the seen classes will likely allow the model to generalize better to the unseen classes.

We emphasize that this is different from the standard generalization notion between training and test data. In that scenario, the network is evaluated on how well it performs on held-out data points belonging to the same classes as those present in the training data. This can be framed as ``weak generalization'', and may be interpreted geometrically as testing the network on novel points sampled from the same manifold $\mathcal{M} \in \mathbb{R}^d$, with $d \leq m$, where $m$ is the dimension of the input space. The basic assumption is that, if the network is presented with sufficiently varied inputs, it should be able to ``interpolate'' between those to perform well on unseen inputs from the same distribution. The degree to which this will be successful is a matter of how much the network can avoid overfitting, and techniques such as weight regularization \cite{louizos2018learning}, dropout, and optimizing batch size \cite{keskar2017on} are commonly used to help in that regard, although if has been shown that some of these are not sufficient to explain why large networks generalize in practice \cite{zhang2021understanding}. It has also been suggested that this type of generalization could be related to the presence of flatness of local minima in the loss function landscape \cite{dinh2017sharp}.




\subsection{Related Work}
Problems related to an \emph{open set} notion of generalization have been previously investigated in the deep learning literature.
Domain adaptation \cite{ben2006analysis, wang2018deep} considers a change in distribution/domain of inputs (e.g., going from photos to  paintings), but maintaining the same classes (or subset thereof, in the case of partial domain adaptation (PDA) \cite{angeletti2018adaptive, bucci2019tackling}). The key goal is to learn domain-invariant features for each class that translate well across domains, using labeled data from both the `source' and `target' domains. Unsupervised domain adaptation (UDA) \cite{ben2010theory, kang2019contrastive, bousmalis2017unsupervised} is a variant of this problem where the target domain is unlabeled; this is closer to our scenario, except we take it a step further in that even the classes are not the same. Thus, we cannot use the strategy of `matching' same-class data points from one domain into the other.

Out-of-distribution (OOD) or out-of-sample detection \cite{hendrycks2016baseline}---of which anomaly detection, outlier detection, novelty detection, and open-set recognition are special cases \cite{yang2021oodsurvey}---is related to our setting because the novel input samples come from classes unrelated to those seen during training. However, the task is to use binary classification to distinguish between seen data (training + test sets) and unseen data (OOD samples). Out-of-distribution generalization \cite{liu2021towards} addresses the case where the test-set distribution may diverge from that of the training set, so it can be also seen as domain adaption.

\begin{figure}[ht]
    \centering
    \includegraphics[width=0.5\textwidth]{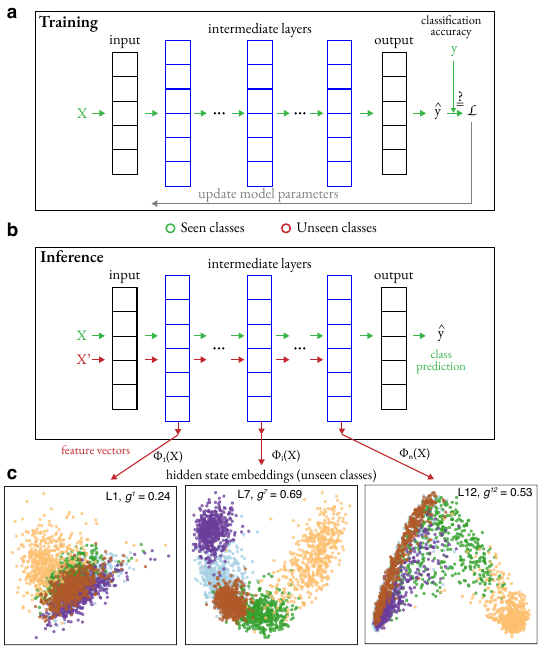}
    \caption{{\bf Schematic of our method of assessing generalizability through out-of-sample embeddings using intermediate layers.}\ \  (\textbf{a}) During training, only a subset of all classes in the dataset are utilized (\textit{seen} classes, in green). (\textbf{b}) At the inference stage, one may use the model to classify novel points from the seen classes (green) or to extract intermediate feature vectors $\Phi_i(X)$ from an intermediate layer $i$ to assess the degree of separability of the unseen classes, as measured by our index $g^{i}$ (eq. 1). (\textbf{c}) Embeddings from different hidden states of the Vision Transformer (ViT) network produce widely varying results. Color labels indicate ground truth: clustered unseen classes indicate better generalization ($g$).\\}
    \label{fig:vert}
\end{figure}

The notion of taking advantage of latent feature representations of the data is particularly important in the field of zero-shot learning \cite{xian2018zero, pourpanah2022review}. Traditionally, the goal is to infer the class of images from unseen classes based on some form of annotation: semantic attributes \cite{farhadi2009describing, maniyar2020zero, han2021contrastive}, word vector \cite{frome2013devise, socher2013zero}, or a short text description/caption \cite{reed2016learning} that describes them. Seen and unseen classes are related in a so-called `semantic space', where the knowledge from seen classes can be transferred to unseen classes by means of the annotations. 
In one such approach, the model learns a joint embedding space onto which both the semantic vectors and the visual feature vectors can be projected \cite{yang2014unified,lei2015predicting}. An alternative is to learn a mapping from one to the other \cite{socher2013zero, frome2013devise}). Novel images can be classified by finding the class that is nearest to it in the semantic space.

In one-shot and few-shot learning \cite{brown2020language, chen2019closer, dhillon2019baseline}, a model is given a single, or few, labeled examples of an unseen class along with an unlabeled example. The model predicts the label based on how similar it is to the novel, labeled examples. To achieve this, the model is expected to have a powerful feature-level representation, but is still reliant on labeled classes.

Closest to our scenario is the notion of novel category discovery \cite{han2019learning},  in which the challenge is to infer novel classes in unlabelled data points using a labelled subset of the data. Furthermore, the task of fully labelling novel data that may include both seen and unseen classes has been framed as `generalized category discovery' \citep{vaze2022generalized,vaze2024no}. Such studies, however, have utilized specific architectures (e.g., deep embedded clustering) which incorporate clustering as a later stage of the model; this requires the model to be actually optimized to produce good clusters (using labelled unseen classes). In contrast, we here investigate how well can some of the most popular pre-trained models generalize to unseen categories directly after being fine-tuned on a related domain (i.e., without being specifically trained to enhance generalization).

Therefore our task may also be framed in terms of a generalized zero-shot setting \cite{han2021contrastive}, in which the goal is to correctly organize samples from both seen and unseen labels, except that we do not employ semantic information beyond the visual features already present in the input images. Instead, we use multiple unlabeled points as context in order to infer class structure. In contrast, a large language model (LLM) performs zero-shot inference by receiving additional context about a new class(es) in the same prompt (`zero-shot prompting') \cite{wei2021finetuned, kojima2022large}. For example: an autoregressive language model may respond correctly to the prompt: \textit{``Classify the sentiment of following sentence into positive or negative: `I enjoyed this paper.' Sentiment: ''} even if it had not been trained to perform sentiment analysis. In our specific setting of image classification, a context is given in the form of many additional inputs coming from the unseen classes.

We work with a purely visual setting, as in \cite{chen2019closer, dhillon2019baseline}. The idea is to utilize a minimalist approach in order to avoid confounds from the method of embedding semantic information and of relating visual to semantic features. This paradigm is relevant for the common real-world, open set scenario, in which many images are available without annotation. Ultimately, we aim to test whether the learned feature vectors are sufficient to support zero-shot learning (in the sense that we do not have labels for the unseen classes) or few-shot learning (in the sense that we need multiple examples to assess proximity between data points).

Three approaches are used to evaluate the success of such predictions (see details in Methodology): $K$-means (which assumes that classes should form Gaussian-like clusters); $k$-nearest neighbors (assumes that samples should be closer to the $k$ closest samples from the same unseen class than those from other classes); and a linear probe classifier to  directly measure how separable the unseen classes are in the latent space---it addresses the practical case where one is aware of the novelty of the classes being used for during inference (and therefore can label the outputs), but fine-tuning the model is infeasible. Thus, the linear probe emulates an $n$-way few-shot learning, where labeled unseen classes can be seen as the \textit{support set}. This supervised technique also resembles transductive few-shot learning \cite{nichol2018first, dhillon2019baseline, lazarou2021iterative}, in which all unseen examples are classified at once.


\section{Methodology}

\subsection{Models and data}

To test our approach, we fine-tuned six pretrained networks for visual classification : ViT-base (ViT) \cite{vit}, Swin Transformer (Swin)\citep{swin}, Pyramid ViT (PViT) \cite{pvt}, CvT-21 (CvT) \cite{cvt}, PoolFormer-S12 (PF) \cite{poolformer}, ConvNeXt V2 (CNV2) \cite{convnext2}. Our goal was to experiment with a representative set of state-of-the-art models: we used four transformers (but PoolFormer does not use attention layers and CvT uses convolutional layers) and one fully convolutional network (ConvNeXt V2).
We used two different datasets for fine tuning: the CIFAR-100 natural scenes dataset \cite{Krizhevsky09learningmultiple}, which classifies images by their content, and a Chinese calligraphy dataset \citep{dataset}, which classifies grayscale images of drawn characters by the artist that drew them. For each dataset, we sampled 15 classes to be seen only during training (the \textit{seen} classes) and 5 to be withheld for assessing generalization (the \textit{unseen} classes). \footnote{Because all the networks were pre-trained on ImageNet-1k, a dataset that shares considerable overlap in classes with CIFAR-100, we needed to find classes in CIFAR-100 that were not present in ImageNet-1k. Fortunately, we observed that ImageNet-1k does not have classes for flower species but CIFAR-100 does, so we were able to use ‘sunflower’, ‘tulip’, ‘orchid’, ‘poppy’, and ‘rose’ as the unseen classes. There are no Chinese calligraphy images in ImageNet-1k, so this was not a concern for the calligraphy dataset.} 
Our approach to fine-tune the models only on the seen classes is in contrast with other works investigating few-shot learning where the model is fine-tuned on the support (unseen) set \cite[e.g.,][]{dhillon2019baseline}.

The networks were fine-tuned using PyTorch and the $\texttt{transformers}$ package for 500 epochs on the seen classes using the following hyperparameters: learning rate $2\mathrm{e}{-4}$, batch size 72; AdamW optimizer.

\subsection{Category generalization}

To assess generalization to unseen classes, we used the intuition that intermediate embeddings of examples from learned classes should form separable clusters. Thus, we created a \emph{generalization index}, $g$, that measures the degree of separability of examples $\{x\}$ within latent space embeddings $\{\Phi_{i}(x)\}$ where $i$ indexes the intermediate layer providing the embedding.

For a given network, generalization can be assessed in terms of the quality of a $K$-means cluster assignment (using Euclidean distance and $K$ equal to the number of unseen classes) computed on the embedding of unseen examples when compared to the ground truth. This comparison can be done by first computing the normalized mutual information (NMI) \citep{danon2005comparing} between the two assignments:
\begin{equation}\label{eq:nmi}
\mathrm{NMI}(A,B) = \frac{-2 \sum_{i=1}^{c} \sum_{j=1}^{c} N_{ij} \log\left(\frac{N_{ij}N}{N_{i\cdot}N_{\cdot j}}\right)}
{\sum_{i=1}^{c}N_{i\cdot}\log\left(\frac{N_{i\cdot}}{N}\right) + \sum_{j=1}^{c}N_{\cdot j}\log\left(\frac{N_{\cdot j}}{N}\right)}
\end{equation}
where $c$ is the number of classes and $\mathbf{N}$ is a confusion matrix with entries $N_{ij}$ corresponding to the number of points in the class $i$ that appear in the cluster $j$ found by $K$-means; $N_{i\cdot}$ denotes the sum over a row, $N_{\cdot j}$ a sum over a column, and $N$ the total number of points.

Then, $g^{i}_{\text{unseen}}$ is computed as
\begin{equation}\label{eq:g_eq}
g^{i}_{\text{unseen}} = \left\{\mathrm{NMI}\left ( \mathcal{C}^{\Phi_i}_{\mathrm{unseen}}, \mathcal{C}^{\star}\right)\right\}
\end{equation}
where $i$ indexes the intermediate layers, $\mathcal{C}^{\Phi_i}_{\mathrm{unseen}}$ denotes the $K$-means cluster assignments of the unseen examples embedded in $\Phi_i$, and $\mathcal{C}^\star$ denotes the images’ true labels.

We can also define the overall generalization power of a network by using the layer that generalizes the best:
\begin{equation}
    g = \max_{i} g^{i}.
\end{equation}

\begin{figure}[ht]
    \centering
\includegraphics[width=0.5\textwidth]{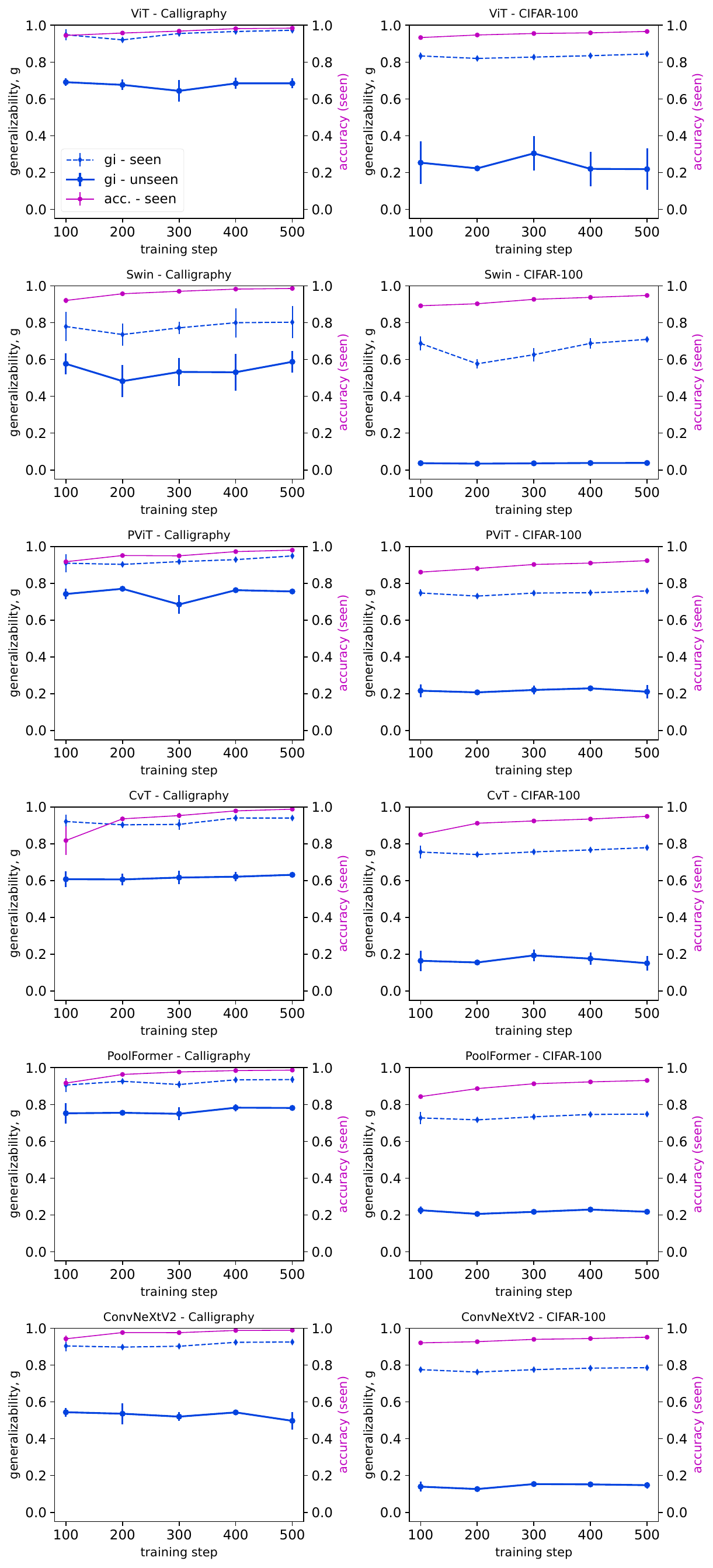}
    \caption{{\bf Generalizability to unseen classes varies across architectures}, even though accuracy increases roughly monotonically across training epochs. We plot $g_{\text{unseen}}=\mathrm{max}_{i} (g^{i})$ and show generalizability to seen classes ($g_{\text{seen}}$). $g_{\text{seen}}$ always dominates $g_{\text{unseen}}$ (as expected). While one might assume that high classification accuracy implies the model has learned a representation of its complete domain, these plots suggest that it is fitting well (or \textit{overfitting}) only the sub-domain sampled by the training data. (Error bars denote std. dev.)\\
    }
    \label{fig:acc-gs}
\end{figure}
\begin{figure}[ht]
    \centering
    \includegraphics[width=0.45\textwidth]{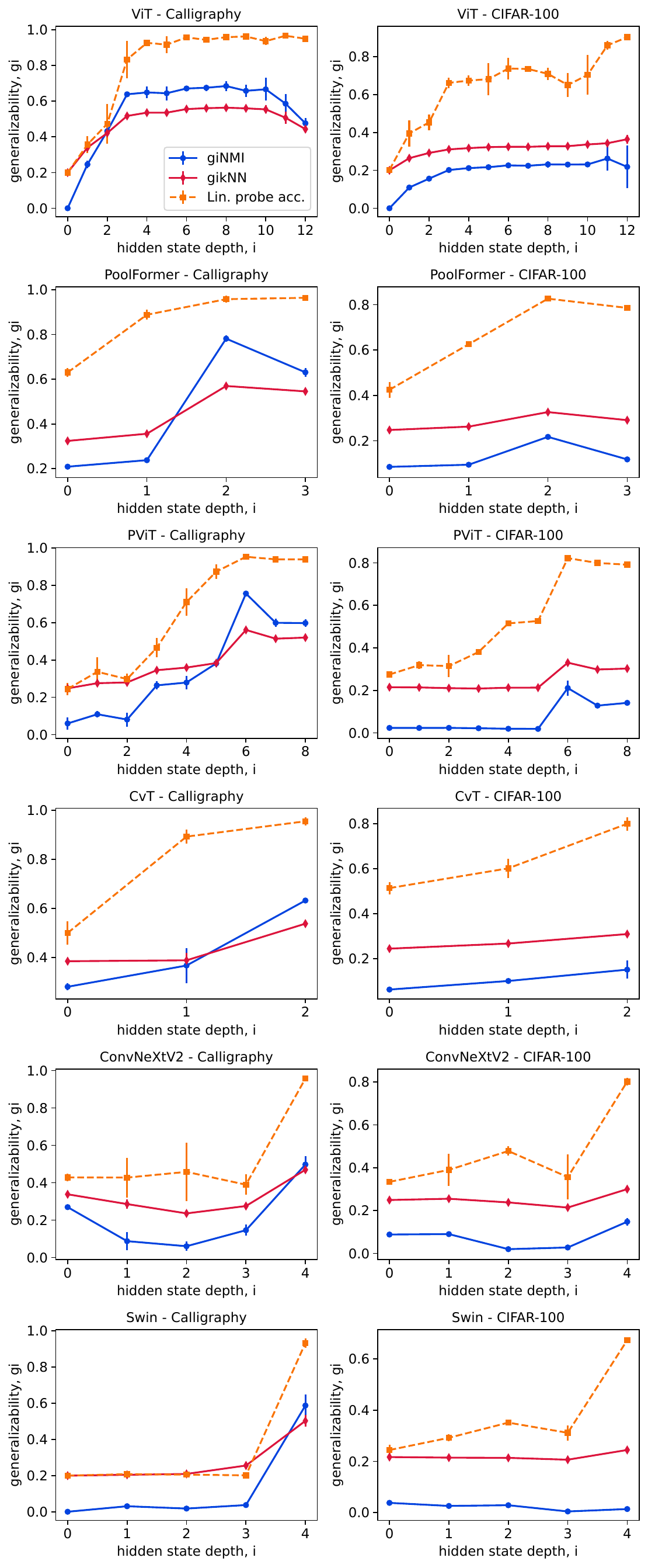}
    \caption{ {\bf Generalizability varies differently across depth in different networks.} For ViT, maximum values of $g^i$ are achieved in early layers (top); for Swin only at the final stage (bottom).
    $g^i$ is not monotonic with depth and, for many models (ViT, PViT, and PoolFormer), the best generalization resides at intermediate layers. This holds true across datasets and metrics. (Note, especially, the agreement between the two unsupervised methods, NMI and kNN). In most cases, all metrics identified the same layer as the most generalizable to unseen classes. We conclude that, since the $g^i$ curves are qualitatively similar across datasets, the patterns observed follow from the networks' architecture, and are not specific to a dataset.}
    \label{fig:metrics}
\end{figure}

\FloatBarrier

To compare the separability of the unseen class embeddings $g^{i}_{\text{unseen}}$ to those of the seen class embeddings, we also compute $g^{i}_{\text{seen}}$ analogously by obtaining $K$-means cluster labels for only the seen examples and comparing those to the ground truth.

To validate our choice of metric, we compared $g$ to another unsupervised metric based on $k$-nearest neighbors ($g_{\text{kNN}}$), as well as a supervised metric based on linear probes ($g_{\text{LPr}}$).

The use of a $k$-nearest neighbors-based metric relies on the intuitive notion that nearest neighbors should belong to the same class.  For each data point, we computed its $k$-nearest neighbors (kNN) in an embedding, using Euclidean distances and setting $k$ to the number of examples in each class. To compute $g^{i}_{\text{kNN}}$, we used the mean, over all data points, of the fraction of a data point $x$'s $k$ nearest neighbors belonging to the same class as $x$. This alternative to the NMI-based $g^{i}$ was used because the number of unseen classes set as $K$ in $K$-means could mis-estimate the number of clusters actually present in the embedding, in which case NMI would not be a good estimate but kNN could still be.

For the linear probe method, we trained a linear classification head using each intermediate layer's output after showing it a training set of 500 examples from the unseen classes and then testing it on 360 more examples from the unseen classes.

To control for randomness in the training, we fine-tuned and calculated metrics for each model three times using different seeds and computed the average of each result, along with standard deviations.

In summary, we adopt the position that, for successful category generalization, representations should be separable, or clustered (analogously to \citet{han2019learning}), but the clustering assessment should be applicable to intermediate layers and it should be calculable without specifying the semantics of a particular taxonomy. We address this apparent contradiction by defining a measure of clustering that can be assessed ``after the fact,'' and apply it to multiple datasets using multiple algorithms. While the results are thus empirical, consistency across algorithms and problems provides a measure of confidence; the conclusions in this paper reflect this confidence.

\section{Results}
After fine-tuning six networks on two datasets and measuring their generalization performance via several metrics, we found that $g$, i.e. the max $g^{i}_{\text{NMI}}$ across all layers, is always lower on the unseen data, compared to the seen data (as expected). Furthermore, the difference is often quite stark, especially on the CIFAR dataset, as can be seen in Figure~\ref{fig:acc-gs}. A low $g$ means that regardless of classification accuracy, an intermediate-layer based embedding from the network would not be useful unless that particular class had been encountered during training.

Additionally, while training a particular network, a higher classification accuracy did not always lead to better generalization. While our generalizability metrics on the seen classes tend to improve with classification accuracy, generalizability on the unseen classes often plateaus and, in most networks, decreases at least once during training.

Looking at generalizability across all layers---not just the best layer---, there is no universal trend as to which layer will provide the best representation for separating unseen examples; sometimes the last layer is best, but often an earlier layer is better, as can be seen in Figure~\ref{fig:metrics}. It is usually the case, however, that a network’s most generalizable layer identified for one dataset, will be the same for a different dataset. Comparing across datasets, the layer generalization curves are qualitatively similar, indicating that our metric captures an intrinsic aspect of the architecture.

Furthermore, the different metrics tend to agree qualitatively. The $g_{\text{kNN}}$ curves align well with $g_{\text{NMI}}$ for a given architecture, demonstrating that the assumption that the classes should be clustered is reasonable. The linear probe results are likewise similar with regard to the relative performance of each layer. Its values are higher across the board, which is unsurprising since the linear probe is a supervised approach and trained on labeled examples, in contrast to the unsupervised cluster-separation based approach. Overall, the findings of the three metrics agree, reinforcing their conclusions.

The model accuracies (\% of correctly classified examples on the seen classes using validation sets) and results for all three metrics on seen and unseen classes are reported in Table \ref{table:network_comparison_cifar} for the CIFAR-100 dataset, and Table \ref{table:network_comparison_callig} for the calligraphy dataset. Boldface denotes the highest values achieved for each metric.

\vspace{0.25cm}

\begin{table}[h!]
\caption{Generalization $g$ of classification networks for unseen and seen classes after fine-tuning on CIFAR-100 dataset.}
\centering
\begin{tabular}{l|cccccc}
\toprule
Network &  ViT & Swin & PViT & CvT & PF & CNV2\\
\hline
accuracy            & {\bf 0.97} & 0.95 & 0.92 & 0.95 & 0.93 & 0.95 \\
\hline
$g_{\text{\tiny NMI,seen}}$ & {\bf 0.84} & 0.71 & 0.76 & 0.78 & 0.75 & 0.79\\
$g_{\text{\tiny NMI,unseen}}$ & {\bf 0.26} & 0.04 & 0.21 & 0.15 & 0.22 & 0.15\\
\hline
$g_{\text{\tiny kNN,seen}}$   & {\bf 0.20} & 0.13 & {\bf 0.20} & {\bf 0.20} & 0.19 & {\bf 0.20}\\
$g_{\text{\tiny kNN,unseen}}$    & {\bf 0.36} & 0.24 & 0.33 & 0.31 & 0.33 & 0.30\\
\hline
$g_{\text{\tiny LPr,seen}}$   & {\bf 0.96} & 0.92 & 0.92 & 0.95 & 0.92 & 0.94\\
$g_{\text{\tiny LPr,unseen}}$   & {\bf 0.90} & 0.67 & 0.82 & 0.80 & 0.83 & 0.80\\ 
\bottomrule
\end{tabular}
\label{table:network_comparison_cifar}
\end{table}

\begin{table}[h!]
\caption{Generalization $g$ of classification networks for unseen and seen classes after fine-tuning on calligraphy dataset.}
\centering
\begin{tabular}{l|cccccc}
\toprule
Network &  ViT & Swin & PViT & CvT & PF & CNV2 \\
\hline
accuracy            & 0.98 & {\bf 0.99} & 0.98 & {\bf 0.99} & {\bf 0.99} & {\bf 0.99} \\
\hline
$g_{\text{\tiny NMI,seen}}$ & {\bf 0.97} & 0.8 & 0.95 & 0.94 & 0.94 & 0.93\\
$g_{\text{\tiny NMI,unseen}}$ & 0.68 & 0.59 & 0.76 & 0.63 & {\bf 0.78} & 0.5\\
\hline
$g_{\text{\tiny kNN,seen}}$   & {\bf 0.38} & 0.34 & {\bf 0.38} & 0.37 & 0.37 & 0.37\\
$g_{\text{\tiny kNN,unseen}}$    & 0.56 & 0.5 & 0.56 & 0.54 & {\bf 0.57} & 0.47\\
\hline
$g_{\text{\tiny LPr,seen}}$   & {\bf 0.99} & 0.95 & 0.98 & {\bf 0.99} & 0.98 & {\bf 0.99}\\
$g_{\text{\tiny LPr,unseen}}$   & {\bf 0.97} & 0.93 & 0.95 & 0.96 & 0.96 & 0.96\\
\bottomrule
\end{tabular}
\label{table:network_comparison_callig}
\end{table}

\begin{figure}[h]
   \centering
   \includegraphics[width=0.4\textwidth]{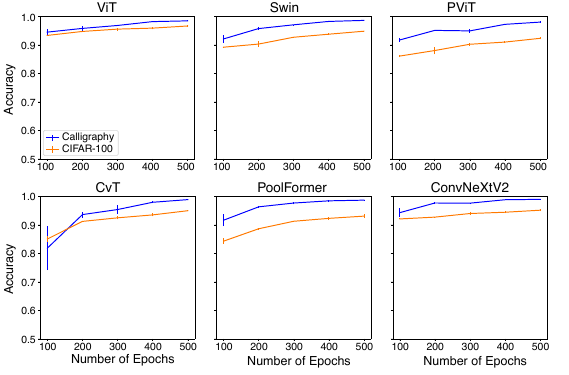}
   \caption{Six popular models were fine-tuned for 500 epochs on the Calligraphy and CIFAR-100 datasets. Each model's classification accuracy after every 100 epochs is shown above. Comparable performance and qualitatively similar accuracy curves were observed for all models (although the CIFAR-100 dataset was more challenging). This is in contrast with generalization power (see Fig.~\ref{fig:acc-gs}).\\ }
   \label{fig:accs}
\end{figure}

\section{Conclusion}

As current models become larger and increasingly expensive to train, due to the cost of manually labeling many images, hardware, and energy consumption, there is a real necessity for developing models that can reliably organize data from related domains in such a way that allows unseen classes to be distinguished (e.g., for few-shot learning).

Intuitively, different architectures are likely to impose different inductive biases, which may or may not help with generalization. First, we confirmed that higher accuracy on a subset of the domain (seen classes) does not imply higher generalizability: although all models reached high classification accuracy after fine-tuning (at least 95\%, see Fig.~\ref{fig:accs}), they achieved widely different generalization powers.

Second, our experiments demonstrated the central role architecture plays: some architectures maximize generalization in shallow layers, while others only generalize at the end. This has obvious implications for pruning and improving model efficiency at inference time. In the case of ViT, for example, less than a third of the full network is needed to achieve the highest levels of generalizability.
We believe that our proposed framework can be used to test architectural modifications and their impact on inferring unseen classes, and thereby guide future architectural design and improvements.

Future work in this area would look at specific ways to improve generalizability through architecture design (e.g., number of layers, layer size, etc.), training paradigms (e.g., contrastive learning), or regularization techniques (e.g., dropout). Crucially, our method can be used to quantify which of these are actually important.






\begin{ack}
Supported by NIH Grant 1R01EY031059, NSF Grant 1822598.
\end{ack}


\newpage

\bibliography{mybibfile}

\begin{thebibliography}{45}
\providecommand{\natexlab}[1]{#1}
\providecommand{\url}[1]{\texttt{#1}}
\expandafter\ifx\csname urlstyle\endcsname\relax
  \providecommand{\doi}[1]{doi: #1}\else
  \providecommand{\doi}{doi: \begingroup \urlstyle{rm}\Url}\fi

\bibitem[Alain and Bengio(2017)]{alain17understanding}
G.~Alain and Y.~Bengio.
\newblock Understanding intermediate layers using linear classifier probes.
\newblock In \emph{5th International Conference on Learning Representations, {ICLR} 2017, Toulon, France, April 24-26, 2017, Workshop Track Proceedings}. OpenReview.net, 2017.

\bibitem[Angeletti et~al.(2018)Angeletti, Caputo, and Tommasi]{angeletti2018adaptive}
G.~Angeletti, B.~Caputo, and T.~Tommasi.
\newblock Adaptive deep learning through visual domain localization.
\newblock In \emph{2018 IEEE International Conference on Robotics and Automation (ICRA)}, pages 7135--7142. IEEE, 2018.

\bibitem[Ben-David et~al.(2006)Ben-David, Blitzer, Crammer, and Pereira]{ben2006analysis}
S.~Ben-David, J.~Blitzer, K.~Crammer, and F.~Pereira.
\newblock Analysis of representations for domain adaptation.
\newblock \emph{Advances in Neural Information Processing Systems}, 19, 2006.

\bibitem[Ben-David et~al.(2010)Ben-David, Blitzer, Crammer, Kulesza, Pereira, and Vaughan]{ben2010theory}
S.~Ben-David, J.~Blitzer, K.~Crammer, A.~Kulesza, F.~Pereira, and J.~W. Vaughan.
\newblock A theory of learning from different domains.
\newblock \emph{Machine learning}, 79:\penalty0 151--175, 2010.

\bibitem[Bousmalis et~al.(2017)Bousmalis, Silberman, Dohan, Erhan, and Krishnan]{bousmalis2017unsupervised}
K.~Bousmalis, N.~Silberman, D.~Dohan, D.~Erhan, and D.~Krishnan.
\newblock Unsupervised pixel-level domain adaptation with generative adversarial networks.
\newblock In \emph{Proceedings of the IEEE Conference on Computer Vision and Pattern Recognition}, pages 3722--3731, 2017.

\bibitem[Brown et~al.(2020)Brown, Mann, Ryder, Subbiah, Kaplan, Dhariwal, Neelakantan, Shyam, Sastry, Askell, et~al.]{brown2020language}
T.~Brown, B.~Mann, N.~Ryder, M.~Subbiah, J.~D. Kaplan, P.~Dhariwal, A.~Neelakantan, P.~Shyam, G.~Sastry, A.~Askell, et~al.
\newblock Language models are few-shot learners.
\newblock \emph{Advances in neural information processing systems}, 33:\penalty0 1877--1901, 2020.

\bibitem[Bucci et~al.(2019)Bucci, D’Innocente, and Tommasi]{bucci2019tackling}
S.~Bucci, A.~D’Innocente, and T.~Tommasi.
\newblock Tackling partial domain adaptation with self-supervision.
\newblock In \emph{Image Analysis and Processing--ICIAP 2019: 20th International Conference, Trento, Italy, September 9--13, 2019, Proceedings, Part II 20}, pages 70--81. Springer, 2019.

\bibitem[Chen et~al.(2019)Chen, Liu, Kira, Wang, and Huang]{chen2019closer}
W.-Y. Chen, Y.-C. Liu, Z.~Kira, Y.-C.~F. Wang, and J.-B. Huang.
\newblock A closer look at few-shot classification.
\newblock \emph{arXiv preprint arXiv:1904.04232}, 2019.

\bibitem[Danon et~al.(2005)Danon, Diaz-Guilera, Duch, and Arenas]{danon2005comparing}
L.~Danon, A.~Diaz-Guilera, J.~Duch, and A.~Arenas.
\newblock Comparing community structure identification.
\newblock \emph{Journal of Statistical Mechanics: Theory and experiment}, 2005\penalty0 (09):\penalty0 P09008, 2005.

\bibitem[Dhillon et~al.(2019)Dhillon, Chaudhari, Ravichandran, and Soatto]{dhillon2019baseline}
G.~S. Dhillon, P.~Chaudhari, A.~Ravichandran, and S.~Soatto.
\newblock A baseline for few-shot image classification.
\newblock \emph{arXiv preprint arXiv:1909.02729}, 2019.

\bibitem[Dinh et~al.(2017)Dinh, Pascanu, Bengio, and Bengio]{dinh2017sharp}
L.~Dinh, R.~Pascanu, S.~Bengio, and Y.~Bengio.
\newblock Sharp minima can generalize for deep nets.
\newblock In \emph{International Conference on Machine Learning}, pages 1019--1028. PMLR, 2017.

\bibitem[Dosovitskiy et~al.(2020)Dosovitskiy, Beyer, Kolesnikov, Weissenborn, Zhai, Unterthiner, Dehghani, Minderer, Heigold, Gelly, et~al.]{vit}
A.~Dosovitskiy, L.~Beyer, A.~Kolesnikov, D.~Weissenborn, X.~Zhai, T.~Unterthiner, M.~Dehghani, M.~Minderer, G.~Heigold, S.~Gelly, et~al.
\newblock An image is worth 16x16 words: Transformers for image recognition at scale.
\newblock \emph{arXiv preprint arXiv:2010.11929}, 2020.

\bibitem[Farhadi et~al.(2009)Farhadi, Endres, Hoiem, and Forsyth]{farhadi2009describing}
A.~Farhadi, I.~Endres, D.~Hoiem, and D.~Forsyth.
\newblock Describing objects by their attributes.
\newblock In \emph{2009 IEEE conference on computer vision and pattern recognition}, pages 1778--1785. IEEE, 2009.

\bibitem[Frome et~al.(2013)Frome, Corrado, Shlens, Bengio, Dean, Ranzato, and Mikolov]{frome2013devise}
A.~Frome, G.~S. Corrado, J.~Shlens, S.~Bengio, J.~Dean, M.~Ranzato, and T.~Mikolov.
\newblock Devise: A deep visual-semantic embedding model.
\newblock \emph{Advances in Neural Information Processing Systems}, 26, 2013.

\bibitem[Han et~al.(2019)Han, Vedaldi, and Zisserman]{han2019learning}
K.~Han, A.~Vedaldi, and A.~Zisserman.
\newblock Learning to discover novel visual categories via deep transfer clustering.
\newblock In \emph{Proceedings of the IEEE/CVF International Conference on Computer Vision}, pages 8401--8409, 2019.

\bibitem[Han et~al.(2021)Han, Fu, Chen, and Yang]{han2021contrastive}
Z.~Han, Z.~Fu, S.~Chen, and J.~Yang.
\newblock Contrastive embedding for generalized zero-shot learning.
\newblock In \emph{Proceedings of the IEEE/CVF conference on computer vision and pattern recognition}, pages 2371--2381, 2021.

\bibitem[Hendrycks and Gimpel(2016)]{hendrycks2016baseline}
D.~Hendrycks and K.~Gimpel.
\newblock A baseline for detecting misclassified and out-of-distribution examples in neural networks.
\newblock \emph{arXiv preprint arXiv:1610.02136}, 2016.

\bibitem[Kang et~al.(2019)Kang, Jiang, Yang, and Hauptmann]{kang2019contrastive}
G.~Kang, L.~Jiang, Y.~Yang, and A.~G. Hauptmann.
\newblock Contrastive adaptation network for unsupervised domain adaptation.
\newblock In \emph{Proceedings of the IEEE/CVF Conference on Computer Vision and Pattern Recognition}, pages 4893--4902, 2019.

\bibitem[Keskar et~al.(2017)Keskar, Mudigere, Nocedal, Smelyanskiy, and Tang]{keskar2017on}
N.~S. Keskar, D.~Mudigere, J.~Nocedal, M.~Smelyanskiy, and P.~T.~P. Tang.
\newblock On large-batch training for deep learning: Generalization gap and sharp minima.
\newblock In \emph{International Conference on Learning Representations}, 2017.
\newblock URL \url{https://openreview.net/forum?id=H1oyRlYgg}.

\bibitem[Kojima et~al.(2022)Kojima, Gu, Reid, Matsuo, and Iwasawa]{kojima2022large}
T.~Kojima, S.~S. Gu, M.~Reid, Y.~Matsuo, and Y.~Iwasawa.
\newblock Large language models are zero-shot reasoners.
\newblock \emph{Advances in neural information processing systems}, 35:\penalty0 22199--22213, 2022.

\bibitem[Krizhevsky(2009)]{Krizhevsky09learningmultiple}
A.~Krizhevsky.
\newblock Learning multiple layers of features from tiny images.
\newblock Technical report, 2009.

\bibitem[Lazarou et~al.(2021)Lazarou, Stathaki, and Avrithis]{lazarou2021iterative}
M.~Lazarou, T.~Stathaki, and Y.~Avrithis.
\newblock Iterative label cleaning for transductive and semi-supervised few-shot learning.
\newblock In \emph{Proceedings of the IEEE/CVF International Conference on Computer Vision}, pages 8751--8760, 2021.

\bibitem[Lei~Ba et~al.(2015)Lei~Ba, Swersky, Fidler, et~al.]{lei2015predicting}
J.~Lei~Ba, K.~Swersky, S.~Fidler, et~al.
\newblock Predicting deep zero-shot convolutional neural networks using textual descriptions.
\newblock In \emph{Proceedings of the IEEE international Conference on Computer Vision}, pages 4247--4255, 2015.

\bibitem[Liu et~al.(2021{\natexlab{a}})Liu, Shen, He, Zhang, Xu, Yu, and Cui]{liu2021towards}
J.~Liu, Z.~Shen, Y.~He, X.~Zhang, R.~Xu, H.~Yu, and P.~Cui.
\newblock Towards out-of-distribution generalization: A survey.
\newblock \emph{arXiv preprint arXiv:2108.13624}, 2021{\natexlab{a}}.

\bibitem[Liu et~al.(2021{\natexlab{b}})Liu, Lin, Cao, Hu, Wei, Zhang, Lin, and Guo]{swin}
Z.~Liu, Y.~Lin, Y.~Cao, H.~Hu, Y.~Wei, Z.~Zhang, S.~Lin, and B.~Guo.
\newblock Swin transformer: Hierarchical vision transformer using shifted windows.
\newblock In \emph{Proceedings of the IEEE/CVF International Conference on Computer Vision}, pages 10012--10022, 2021{\natexlab{b}}.

\bibitem[Louizos et~al.(2018)Louizos, Welling, and Kingma]{louizos2018learning}
C.~Louizos, M.~Welling, and D.~P. Kingma.
\newblock Learning sparse neural networks through $l_0$ regularization.
\newblock In \emph{International Conference on Learning Representations}, 2018.
\newblock URL \url{https://openreview.net/forum?id=H1Y8hhg0b}.

\bibitem[Maniyar et~al.(2020)Maniyar, Deshmukh, Dogan, Balasubramanian, et~al.]{maniyar2020zero}
U.~Maniyar, A.~A. Deshmukh, U.~Dogan, V.~N. Balasubramanian, et~al.
\newblock Zero shot domain generalization.
\newblock \emph{arXiv preprint arXiv:2008.07443}, 2020.

\bibitem[Nichol et~al.(2018)Nichol, Achiam, and Schulman]{nichol2018first}
A.~Nichol, J.~Achiam, and J.~Schulman.
\newblock On first-order meta-learning algorithms.
\newblock \emph{arXiv preprint arXiv:1803.02999}, 2018.

\bibitem[Pourpanah et~al.(2022)Pourpanah, Abdar, Luo, Zhou, Wang, Lim, Wang, and Wu]{pourpanah2022review}
F.~Pourpanah, M.~Abdar, Y.~Luo, X.~Zhou, R.~Wang, C.~P. Lim, X.-Z. Wang, and Q.~J. Wu.
\newblock A review of generalized zero-shot learning methods.
\newblock \emph{IEEE transactions on pattern analysis and machine intelligence}, 45\penalty0 (4):\penalty0 4051--4070, 2022.

\bibitem[Reed et~al.(2016)Reed, Akata, Lee, and Schiele]{reed2016learning}
S.~Reed, Z.~Akata, H.~Lee, and B.~Schiele.
\newblock Learning deep representations of fine-grained visual descriptions.
\newblock In \emph{Proceedings of the IEEE Conference on Computer Vision and Pattern Recognition}, pages 49--58, 2016.

\bibitem[Simonyan and Zisserman(2014)]{simonyan2014very}
K.~Simonyan and A.~Zisserman.
\newblock Very deep convolutional networks for large-scale image recognition.
\newblock \emph{arXiv preprint arXiv:1409.1556}, 2014.

\bibitem[Socher et~al.(2013)Socher, Ganjoo, Manning, and Ng]{socher2013zero}
R.~Socher, M.~Ganjoo, C.~D. Manning, and A.~Ng.
\newblock Zero-shot learning through cross-modal transfer.
\newblock \emph{Advances in Neural Information Processing Systems}, 26, 2013.

\bibitem[Vaze et~al.(2022)Vaze, Han, Vedaldi, and Zisserman]{vaze2022generalized}
S.~Vaze, K.~Han, A.~Vedaldi, and A.~Zisserman.
\newblock Generalized category discovery.
\newblock In \emph{Proceedings of the IEEE/CVF Conference on Computer Vision and Pattern Recognition}, pages 7492--7501, 2022.

\bibitem[Vaze et~al.(2024)Vaze, Vedaldi, and Zisserman]{vaze2024no}
S.~Vaze, A.~Vedaldi, and A.~Zisserman.
\newblock No representation rules them all in category discovery.
\newblock \emph{Advances in Neural Information Processing Systems}, 36, 2024.

\bibitem[Wang and Deng(2018)]{wang2018deep}
M.~Wang and W.~Deng.
\newblock Deep visual domain adaptation: A survey.
\newblock \emph{Neurocomputing}, 312:\penalty0 135--153, 2018.

\bibitem[Wang et~al.(2021)Wang, Xie, Li, Fan, Song, Liang, Lu, Luo, and Shao]{pvt}
W.~Wang, E.~Xie, X.~Li, D.-P. Fan, K.~Song, D.~Liang, T.~Lu, P.~Luo, and L.~Shao.
\newblock Pyramid vision transformer: A versatile backbone for dense prediction without convolutions.
\newblock In \emph{Proceedings of the IEEE/CVF International Conference on Computer Vision}, pages 568--578, 2021.

\bibitem[Wang(2020)]{dataset}
Y.~Wang.
\newblock Chinese calligraphy styles by calligraphers.
\newblock \url{https://www.kaggle.com/datasets/yuanhaowang486/chinese-calligraphy-styles-by-calligraphers}, 2020.

\bibitem[Wei et~al.(2021)Wei, Bosma, Zhao, Guu, Yu, Lester, Du, Dai, and Le]{wei2021finetuned}
J.~Wei, M.~Bosma, V.~Y. Zhao, K.~Guu, A.~W. Yu, B.~Lester, N.~Du, A.~M. Dai, and Q.~V. Le.
\newblock Finetuned language models are zero-shot learners.
\newblock \emph{arXiv preprint arXiv:2109.01652}, 2021.

\bibitem[Woo et~al.(2023)Woo, Debnath, Hu, Chen, Liu, Kweon, and Xie]{convnext2}
S.~Woo, S.~Debnath, R.~Hu, X.~Chen, Z.~Liu, I.~S. Kweon, and S.~Xie.
\newblock {ConvNeXt V2}: {C}o-designing and scaling convnets with masked autoencoders.
\newblock In \emph{Proceedings of the IEEE/CVF Conference on Computer Vision and Pattern Recognition}, pages 16133--16142, 2023.

\bibitem[Wu et~al.(2021)Wu, Xiao, Codella, Liu, Dai, Yuan, and Zhang]{cvt}
H.~Wu, B.~Xiao, N.~Codella, M.~Liu, X.~Dai, L.~Yuan, and L.~Zhang.
\newblock Cvt: Introducing convolutions to vision transformers.
\newblock In \emph{Proceedings of the IEEE/CVF International Conference on Computer Vision}, pages 22--31, 2021.

\bibitem[Xian et~al.(2018)Xian, Lampert, Schiele, and Akata]{xian2018zero}
Y.~Xian, C.~H. Lampert, B.~Schiele, and Z.~Akata.
\newblock Zero-shot learning -- a comprehensive evaluation of the good, the bad and the ugly.
\newblock \emph{IEEE Transactions on Pattern Analysis and Machine Intelligence}, 41\penalty0 (9):\penalty0 2251--2265, 2018.

\bibitem[Yang et~al.(2021)Yang, Zhou, Li, and Liu]{yang2021oodsurvey}
J.~Yang, K.~Zhou, Y.~Li, and Z.~Liu.
\newblock Generalized out-of-distribution detection: A survey.
\newblock \emph{arXiv preprint arXiv:2110.11334}, 2021.

\bibitem[Yang and Hospedales(2014)]{yang2014unified}
Y.~Yang and T.~M. Hospedales.
\newblock A unified perspective on multi-domain and multi-task learning.
\newblock \emph{arXiv preprint arXiv:1412.7489}, 2014.

\bibitem[Yu et~al.(2022)Yu, Luo, Zhou, Si, Zhou, Wang, Feng, and Yan]{poolformer}
W.~Yu, M.~Luo, P.~Zhou, C.~Si, Y.~Zhou, X.~Wang, J.~Feng, and S.~Yan.
\newblock Metaformer is actually what you need for vision.
\newblock In \emph{Proceedings of the IEEE/CVF Conference on Computer Vision and Pattern Recognition}, pages 10819--10829, 2022.

\bibitem[Zhang et~al.(2021)Zhang, Bengio, Hardt, Recht, and Vinyals]{zhang2021understanding}
C.~Zhang, S.~Bengio, M.~Hardt, B.~Recht, and O.~Vinyals.
\newblock Understanding deep learning (still) requires rethinking generalization.
\newblock \emph{Communications of the ACM}, 64\penalty0 (3):\penalty0 107--115, 2021.

\end{thebibliography}

\end{document}